\documentclass[12pt]{article}

\usepackage{amsmath}
\usepackage{amssymb}
\usepackage{setspace,enumitem,soul,url}

\usepackage{authblk}
\author[1]{Luis Sanguiao Sande}
\author[2,3,4]{Li-Chun Zhang}
\affil[1]{\em \small Instituto Nacional de Estad\'\i stica (luis.sanguiao.sande@ine.es)}
\affil[2]{\em \small University of Southampton (email: L.Zhang@soton.ac.uk)}
\affil[3]{\em \small Statistisk sentralbyraa}
\affil[4]{\em \small Universitetet i Oslo}
\title{\Large Design-unbiased statistical learning in survey sampling}
\date{}

\begin{document}

\maketitle

\vspace{-8mm}
\emph{Abstract:} Design-consistent model-assisted estimation has become the standard practice in survey sampling. However, a general theory is lacking so far, which allows one to incorporate modern machine-learning techniques that can lead to potentially much more powerful assisting models. We propose a subsampling Rao-Blackwell method, and develop a statistical learning theory for exactly design-unbiased estimation with the help of linear or non-linear prediction models. Our approach makes use of classic ideas from Statistical Science as well as the rapidly growing field of Machine Learning. Provided rich auxiliary information, it can yield considerable efficiency gains over standard linear model-assisted methods, while ensuring valid estimation for the given target population, which is robust against potential mis-specifications of the assisting model at the individual level. 

\bigskip \noindent
\emph{Keywords:} Rao-Blackwellisation, bagging, $pq$-unbiasedness, stability conditions

\section{Introduction}

Approximately design-unbiased model-assisted estimation is not new. It has become the standard practice in survey sampling, following many  influential works such as S\"{a}rndal et al. (1992), Deville and S\"{a}rndal (1992). However, there lacks so far a theory, which allows one to generally incorporate the many common machine-learning (ML) techniques. For instance, according to Breit and Opsomer (2017, p. 203), they ``are not aware of direct uses of random forests in a model-assisted survey estimator''. Since modern ML techniques can often generate more flexible and powerful prediction models, when rich auxiliary feature data are available, the potentials are worth exploring, in any situation where the practical advantages of linear weighting are not essential compared to the efficiency gains that can be achieved by alternative non-linear ML techniques. 

We propose a \emph{subsampling Rao-Blackwell (SRB)} method, which enables \emph{exactly} design-unbiased estimation with the help of linear or non-linear prediction models. Monte Carlo (MC) versions of the proposed method can be used in cases where exact RB method is computationally too costly. The MC-SRB method is still exactly design-unbiased, despite it is somewhat less efficient due to the additional MC error. In practice, though, one can easily balance between the numerical efficiency of the MC-SRB method against the statistical efficiency of the corresponding exact RB method. 

The SRB method makes use of three classic ideas from Statistical Science and Machine Learning. On the one hand, the training-test split of the sample of observations in ML generates \emph{errors} in the test set rather than \emph{residuals}, conditional on the training dataset, which as we shall explain is the key to achieving exact design-unbiasedness. For model-assisted survey estimation we use this idea to remove the finite-sample bias.  On the other hand, Rao-Blackwellisation (Rao, 1945; Blackwell, 1947) and model-assisted estimation (Cassel et al., 1976) are powerful ideas in Statistics and survey sampling, which we apply to ML techniques to obtain design-unbiased survey estimators at the population level.

We shall refer to the amalgamation as \emph{statistical learning}, since the term model-assisted estimation is entrenched with the property of approximate design-unbiasedness (e.g. S\"{a}rndal 2010;  Breit and Opsomer, 2017), whereas the focus of population-level estimation and associated variance estimation is unusual in the ML literature. 

In applications one needs to ensure design-consistency of the proposed SRB method, in addition to exact design-unbiasedness.  The property can readily be established for parametric or many semi-parametric assisting models. But the conditions required for non-parametric algorithmic ML prediction models have so far eluded a treatment in the literature. Indeed, this has been a main reason preventing the incorporation of such ML techniques in model-assisted estimation from survey sampling. We shall develop general stability conditions for design-consistency under both simple random sampling and arbitrary unequal probability sampling designs. 

For the first time, design-unbiased model-assisted estimation can thereby be achieved generally in survey sampling. Wherever rich feature data are available, the approach of statistical learning developed in this paper enables one to adopt suitable ML techniques, which can make much more efficient use of the available auxiliary information. 

The rest of the paper is organised as follows. In Section \ref{linear}, we describe the SRB method that uses an assisting linear model. The underlying ideas of design-unbiased statistical learning are explained, as well as the differences to the standard model-assisted generalised regression estimation. Some basic methods of variance estimation are outlined, where a novel jackknife variance estimator is developed for the SRB method. We move on to non-linear ML techniques in Section \eqref{non-linear}. The similarity and difference to the bootstrap aggregating (Breiman, 1996b) approach are explored. Moreover, we investigate and prove the stability conditions for design-consistency of SRB method that uses non-parametric algorithmic prediction models. Two simulation studies are presented in Section \ref{simulation}, which illustrate the potential gains of the proposed unbiased statistical learning approach, compared to standard linear model-assisted or model-based approaches. A brief summary and topics for future research will be given in Section \ref{summary}.

\section{Unbiased linear estimation} \label{linear}

In this section we consider unbiased linear estimation in survey sampling, which builds on generalised regression (GREG) estimation (S\"{a}rndal et al. 1992). The GREG estimator is the most common estimation method in practical survey sampling. It is consistent under mild regularity conditions, and is often more efficient than exactly unbiased Horvitz-Thompson (HT) estimation (Horvitz and Thompson, 1952). The proposes subsampling Rao-Blackwellisation (SRB) method removes the finite-sample bias of GREG generally, whose relative efficiency is comparable to the standard GREG estimator.

\subsection{Bias correction by subsampling}

Let $s$ be a sample (of size $n$) selected from the population $U$ of size $N$, with probability $p(s)$, where $\sum_{s} p(s) = 1$ over all possible samples under a given sampling design. Let $\pi_i = \mbox{Pr}(i\in s) >0$ be the sample inclusion probability, for each $i\in U$. Let $y_i$ be a survey variable, for $i\in U$, with unknown population total $Y = \sum_{i\in U} y_i$. 

Let the assisting linear model expectation of $y_i$ be given by $\mu(x_i) = x_i^{\top} \beta$, where $x_i$ is the vector of covariates for each $i\in U$. Let $\mu(x_i, s) = x_i^{\top} b$ be the estimator of $\mu(x_i)$, where $b = (\sum_{i\in s} x_i x_i^{\top}/\pi_i)^{-1} \sum_{i\in s} x_i y_i/\pi_i$ is a weighted least squares (WLS) estimator of $\beta$. It is possible to attach additional heteroscedasticity weights in the WLS; but the development below is invariant to such variations, so that it is more convenient to simply ignore it in the notation. Let $X = \sum_{i\in U} x_i$. The GREG estimator of $Y$ is given as
\[
\widehat{Y}_{GR} = X^{\top} b + \sum_{i\in s} (y_i - x_i^{\top} b)/\pi_i 
\] 
While $\widehat{Y}_{GR}$ is design-consistent under mild regularity conditions (e.g. S\"{a}rndal et al. 1992), as $n,N \rightarrow \infty$, it is usually biased given finite sample size $n$, except in special cases such as when $x_i \equiv 1$ and $\pi_i \equiv n/N$, where $\mu(x,s) = \sum_{i\in s} y_i/n = \bar{y}_s$ and $\widehat{Y}_{GR} = N \bar{y}_s$.

To remove the potential finite-sample bias of $\widehat{Y}_{GR}$, consider subsampling of $s_1 \subset s$, with known probability $q(s_1 |s)$, such as SRS with fixed $n_1 = |s_1|$, where $\sum_{s_1} q(s_1 |s) = 1$. The induced probability of selecting $s_1$ from $U$ is given by 
\[
p_1(s_1) = \sum_{s: s_1\subset s} q(s_1 | s) p(s) 
\] 
where $\pi_{1i} = \mbox{Pr}(i\in s_1)$ is the corresponding inclusion probability for $i\in U$. Let $s_2 = s\setminus s_1$ be the complement of $s_1$ in $s$. Let the conditional sampling probability of $s$ given $s_1$ be
\[
p_2(s_2 | s_1) = p(s) q(s_1 | s) /p_1(s_1) 
\]
and let $\pi_{2i} = \mbox{Pr}(i\in s_2 | s_1)$ be the corresponding conditional inclusion probability in $s_2$ for $i\in U\setminus s_1$. Let $\mu(x_i, s_1) = x_i^{\top} b_1$ be the estimate of $\mu(x_i)$ based on the sub-sample $s_1$, where $b_1 = (\sum_{i\in s_1} x_i x_i^{\top}/\pi_{1i})^{-1} \sum_{i\in s_1} x_i y_i/\pi_{1i}$. Let
\begin{equation} \label{Y1}
\widehat{Y}_1 = \sum_{i\in s_1} y_i + \sum_{i\in U\setminus s_1} x_i^{\top} b_1 + \sum_{i\in s_2} (y_i - x_i^{\top} b_1)/\pi_{2i} 
\end{equation}
In other words, it is the sum of $y_i$ in $s_1$ and a difference estimator of the remaining population total based on $s_2$, via $x_i^{\top} b_1$ that does not depend on the observations in $s_2$.

\paragraph{\em Proposition}  The estimator $\widehat{Y}_1$ is conditionally unbiased for $Y$ over $p_2(s_2 |s_1)$ given $s_1$, denoted by $E_2(\widehat{Y}_1 | s_1) = Y$, as well as unconditionally over $p(s)$, denoted by $E_p(\widehat{Y}_1) = Y$.

\noindent
\emph{Proof:} As $\mu(x_i, s_1)$ is fixed for any $i\in U\setminus s_1$ given $s_1$, the last two terms on the right-hand side of \eqref{Y1} is unbiased for $Y - \sum_{i\in s_1} y_i$ given $s_1$. It follows that $\widehat{Y}_1$ is conditionally unbiased for $Y$ given $s_1$; hence, design-unbiased over $p(s)$ unconditionally as well. $\square$

\paragraph{Example: Simple random sampling (SRS)} Suppose SRS without replacement of $s$ from $U$, and $s_1$ from $s$ with fixed size $n_1 = |s_1|$, such that $\pi_{1i} = n_1/N$ and $\pi_{2i} = (n -n_1)/(N-n_1)$. In the special case of $x_i \equiv 1$, $b_1$ is the sample mean in $s_1$, and
\[
\widehat{Y}_1 = n_1 b_1 + (N - n_1) \sum_{i\in s_2} y_i /(n - n_1)
\]
which amounts to using the sample mean in $s_2$ to estimate the population mean outside of the given $s_1$, i.e., instead of using the sample mean in $s$ for the whole population mean. Thus, $\widehat{Y}_1$ achieves unbiasedness generally, but at a cost of increased variance.

\subsection{Rao-Blackwellisation} 

One can reduce the variance of $\widehat{Y}_1$ by the Rao-Blackwell method (Rao, 1945; Blackwell, 1947). The minimal sufficient statistic in the finite population sampling setting is simply $D_s = \{ (i, y_i) : i\in s\}$. Applying the RB method to $\widehat{Y}_1$ by \eqref{Y1} yields $\widehat{Y}_{GR}^*$, which is given by the conditional expectation of $\widehat{Y}_1$ given $D_s$, i.e.
\begin{equation} \label{RB-GR}
\widehat{Y}_{GR}^* = E_q(\widehat{Y}_1 | D_s) = E_q(\widehat{Y}_1 | s)
\end{equation}
where the expectation is evaluated with respect to $q(s_1 | s)$, and the second expression is leaner as long as one keeps in mind that $\{y_i : i\in U\}$ are treated as fixed constants associated with the distinctive units. 

\paragraph{\em Proposition} The estimator $\widehat{Y}_{GR}^*$ is design-unbiased for $Y$, denoted by $E(\widehat{Y}_{GR}^*) =Y$.

\noindent
\emph{Proof:} By construction, the combined randomisation distribution induced by $p$ and $q$ is the same as that induced by $p_1$ and $p_2$, for any $s_1 \cup s_2 = s$ and $s_1\cap s_2 = \emptyset$. Thus,
\[
E(\widehat{Y}_{GR}^*) = E_p(\widehat{Y}_{GR}^*) = E_p\big( E_q(\widehat{Y}_1 | s) \big) = E_1\big( E_2(\widehat{Y}_1 | s_1) \big) 
= E_1\big( Y\big) =Y  \qquad \square
\]

\bigskip
Next, for the variance of $\widehat{Y}_{GR}^*$ over $p(s)$, i.e. $V(\widehat{Y}_{GR}^*) = V_p(\widehat{Y}_{GR}^*)$, we notice
\begin{align*}
V(\widehat{Y}_1) & = E_p\big( V_q(\widehat{Y}_1 | s) \big) + V_p\big( E_q(\widehat{Y}_1 | s) \big) 
= E_p\big( V_q(\widehat{Y}_1 | s) \big) + V_p(\widehat{Y}_{GR}^*) \\
V(\widehat{Y}_1) & = E_1\big( V_2(\widehat{Y}_1 | s_1) \big) + V_1\big( E_2(\widehat{Y}_1 | s_1) \big) 
= E_1\big( V_2(\widehat{Y}_1 | s_1) \big) 
\end{align*}
since $E_2(\widehat{Y}_1 | s_1) = Y$. Juxtaposing the two expressions of $V(\widehat{Y}_1)$ above, we obtain
\begin{equation} \label{V-RB-GR}
V(\widehat{Y}_{GR}^*) = V_p(\widehat{Y}_{GR}^*) = E_1\big( V_2(\widehat{Y}_1 | s_1) \big) - E_p\big( V_q(\widehat{Y}_1 | s) \big)
\end{equation}
where $E_p\big(V_q(\widehat{Y}_1 | s)\big)$ is the variance reduction compared to $\widehat{Y}_1$.

\paragraph{\em Proposition} Provided unbiased variance estimator $\widehat{V}(\widehat{Y}_1)$ with respect to $p_2(s_2 | s_1)$, i.e. $E_2\big( \widehat{V}(\widehat{Y}_1)\big) = V_2(\widehat{Y}_1 | s_1)$, a design-unbiased variance estimator for $\widehat{Y}_{GR}^*$ is given by
\[
\widehat{V}(\widehat{Y}_{GR}^*) = \widehat{V}(\widehat{Y}_1) - V_q(\widehat{Y}_1 | s)
\]

\noindent
\emph{Proof:} By stipulation, we have $E\big( \widehat{V}(\widehat{Y}_1) \big) = E_1 \Big( E_2\big( \widehat{V}(\widehat{Y}_1) \big) \Big) = E_1\Big( V_2(\widehat{Y}_1 | s_1) \Big)$, which is the first term on the right-hand side of \eqref{V-RB-GR}. The result follows immediately. $\square$

\paragraph{Example: SRS, cont'd} In the special case of $x_i \equiv 1$ and $n_1 = n-1$, we have 
\[
\widehat{Y}_{1(i)} = n_1 \bar{y}_{(i)} + (N - n_1) y_i 
\]
if $s_2 = \{i\}$ and $\bar{y}_{(i)}$ denotes the mean in $s_1 = s\setminus \{ i\}$. The RB estimator follows as
\[
\widehat{Y}_{GR}^* = \frac{1}{n} \sum_{i\in s} \widehat{Y}_{1(i)} 
= \frac{n_1}{n} \sum_{i\in s} \bar{y}_{(i)} + \frac{N}{n}\sum_{i\in s} y_i - \frac{n_1}{n} \sum_{i\in s} y_i = \frac{N}{n}\sum_{i\in s} y_i 
\]
which is the usual unbiased full-sample expansion estimator in this case. The RB method thus recovers the lost efficiency of any $\widehat{Y}_{1(i)}$ on its own.

\bigskip
Let $X_1 = \sum_{i\in s_1} x_i$, and $X_1^c = \sum_{i\not \in s_1} x_i = X - X_1$. To express $\widehat{Y}_{GR}^*$ as a linear combination of $\{ y_i : i\in s\}$, we rewrite $\widehat{Y}_1$ as
\begin{align*}
\widehat{Y}_1 & = \sum_{i\in s_1} y_i + \sum_{i\in s_2} \frac{y_i}{\pi_{2i}} + (X - X_1)^{\top} b_1 - \sum_{i\in s_2} \frac{x_i^{\top} b_1}{\pi_{2i}} \\
& = \sum_{i\in s_1} y_i + \sum_{i\in s_2} \frac{y_i}{\pi_{2i}} + (X_1^c - \widehat{X}_1^c)^{\top} b_1 
= \sum_{i\in s} w_i y_i 
\end{align*}
where  
\[
w_i = \begin{cases} 
1 + (X_1^c - \widehat{X}_1^c)^{\top} \big( \sum_{i\in s_1} x_i x_i^{\top}/\pi_{1i} \big)^{-1} x_i/\pi_{1i} & \text{if } i\in s_1 \\ 
1/\pi_{2i} & \text{if } i\in s_2 
\end{cases}
\]
It follows that the RB estimator \eqref{RB-GR} can be given as a linear estimator
\begin{equation} \label{RB-GR-linear}
\widehat{Y}_{GR}^* = \sum_{i\in s} w_i^* y_i \qquad\text{where}\quad w_i^* = E_q(w_i | s)
\end{equation}
This has an important practical advantage that $\{ w_i^* : i\in s\}$ can be applied to produce numerically consistent cross-tabulation of multiple survey variables of interest.

In the case of SRS of $s_1$ with $n_1 = n-1$, the RB weight $w_i^*$ in \eqref{RB-GR-linear} is the average of $w_i$'s over $n$ possible subsamples $s_1$, for a given unit $i\in s$, where $w_i = 1/\pi_{2i}$ when $s_1$ does not include the unit $i$, otherwise $w_i$ is the corresponding GREG weight for $Y_1^c = \sum_{k\not\in s_1} y_k$, which is different for each of the rest $n-1$ subsamples that includes the unit $i$.

\subsection{Relative efficiency to GREG} 

Let $B = E_p(b)$ and $e_i = y_i - x_i^{\top} B$ for $i\in U$. Expanding the GREG estimator $\widehat{Y}_{GR}$ around $(Y, X, B)$ yields 
\[
\widehat{Y}_{GR} \approx \sum_{i\in s} \frac{e_i}{\pi_i} + B^{\top} X
\]
For $\widehat{Y}_1$, the first two terms on the right-hand side of  \eqref{Y1} becomes $X^{\top} b_1$ if there exists a vector $\lambda$ such that $x_i^{\top} \lambda \equiv 1$, in which case $\widehat{Y}_1$ is a function of $(\widehat{Y}_1^c, \widehat{X}_1^c, b_1)$, i.e.
\[
\widehat{Y}_1 = X^{\top} b_1 + \widehat{Y}_1^c - b_1^{\top} \widehat{X}_1^c = \widehat{Y}_1^c + b_1^{\top} (X - \widehat{X}_1^c) 
\]
where $\widehat{Y}_1^c = \sum_{i\in s_2} y_i/\pi_{2i}$ is conditionally unbiased for $Y_1^c = \sum_{i\not \in s_1} y_i$ given $s_1$, and similarly $\widehat{X}_1^c = \sum_{i\in s_2} x_i/\pi_{2i}$ for $X_1^c = \sum_{i\not \in s_1} x_i$. Let $Y_+^c = E_1(Y_1^c)$ and $X_+^c = E_1(X_1^c)$. We have $E_1(b_1) \approx B$, since $b_1$ and $b$ aim at the same population parameter, especially if $n_1$ is close to $n$. In any case, expanding $\widehat{Y}_1$ around $(Y_+^c, X_+^c, B)$ yields   
\begin{align*}
\widehat{Y}_1 
& \approx \Big( Y_+^c + B^{\top} (X -  X_+^c) \Big) + \Big( (\widehat{Y}_1^c - Y_+^c) + (b_1 -B)^{\top} (X - X_+^c)- B^{\top} (\widehat{X}_1^c - X_+^c) \Big) \\
& = \big( \widehat{Y}_1^c - B^{\top} \widehat{X}_1^c \big) + b_1^{\top} (X - X_+^c) + B^{\top} X_+^c
\end{align*}
and
\[
\widehat{Y}_1^c - B^{\top} \widehat{X}_1^c = \sum_{i\in s_2} e_i/\pi_{2i} = \sum_{i\in s} \delta_{2i}^* e_i/\pi_{2i}
\]
where $\delta_{2i}^* = 1$ if $i\in s_2$ and 0 of $i\in s_1$. Thus, we obtain
\begin{equation} \label{linearRB}
\widehat{Y}_{GR}^* = E_q(\widehat{Y}_1 |s) 
\approx \sum_{i\in s} E_q\big( \frac{\delta_{2i}^*}{\pi_{2i}} |s \big) \pi_i \frac{e_i}{\pi_i} + E_q(b_1 |s)^{\top} (X - X_+^c) + B^{\top} X_+^c
\end{equation}

Notice that $B^{\top} X_+^c$ is a constant. Thus, compared to $\widehat{Y}_{GR}$, the variance of $\widehat{Y}_{GR}^*$ involves that of $E_q(b_1 |s)$ in addition. As $n, N\rightarrow \infty$, the first term on the right-hand side of \eqref{linearRB} is $O_p(N/\sqrt{n})$ provided $\pi_i E_q(\delta_{2i}^*/\pi_{2i}) = O_p(1)$, whereas the second term is $O_p(\sqrt{n})$ if $n_1/n = O(1)$ provided the usual regularity conditions for GREG. As long as the sampling fraction $n/N$ is small, the first term will dominate, in which case the variance of the RB estimator $\widehat{Y}_{GR}^*$ is of the same order as that of the GREG estimator $\widehat{Y}_{GR}$.

\paragraph{Example: SRS, cont'd} Let $n_1 = n-k$, where $k = |s_2|$. We have  
\[
\pi_i \pi_{2i}^{-1} E_q(\delta_{2i}^* |s) = \frac{n}{N} \cdot \frac{N - n_1}{k} \cdot \frac{(n-1)!}{(k-1)! (n-k)!} \cdot \frac{k! (n-k)!}{n!} = 1- \frac{n_1}{N} 
\]
Let $S_e^2$ be the population variance of $\{ e_i : i\in U\}$. The variance of the first-term in \eqref{linearRB} is
\[
V_p\Big( \sum_{i\in s} E_q\big( \frac{\delta_{2i}^*}{\pi_{2i}} |s \big) \pi_i \frac{e_i}{\pi_i} \Big) 
= N^2 \big( 1 - \frac{n}{N} \big) \frac{S_e^2}{n} \big( 1- \frac{n_1}{N} \big)^2 
\]
which is actually smaller than the approximate variance of the GREG estimator under SRS, although the difference will not be noteworthy in practical terms, if the sampling fraction $n/N$ is small, since $1- n/N <1-n_1/N < 1$. Meanwhile, due to the additional variance of $E_q(b_1 |s)$, the estimator $\widehat{Y}_{GR}^*$ by unbiased RB method can possibly have a larger variance than the biased GREG (with general $x_i$). It seems that one should use large $n_1$ if possible, to keep the additional variance due to $E_q(b_1 |s)$ small.

\subsection{Delete-one RB method} \label{LOO-RB-GREG}

The largest possible size of $s_1$ is $n_1 = n-1$. We refer to Rao-Blackwellisation based on SRS of $s_1$ with $n_1 = n-1$ as the \emph{delete-one} (or leave-one-out, LOO) RB method. The conditional sampling design $p_2(s_2 | s_1)$ is not measurable in this case, in that one cannot have an unbiased variance estimator $\widehat{V}(\widehat{Y}_1)$ based on a single observation $y_j$ in $s_2 = \{ j\}$.
For an approximate variance estimator, we reconsider the basic case where $\{ y_1, ..., y_n\}$ form a sample of independent and identically distributed (IID) observations, in order to develop an analogy to the classic jackknife variance estimation (Tukey, 1958). 

Denote by $\theta$ the population mean that is also the expectation of each $y_i$, for $i=1, ..., n$. As before, let $\bar{y}_{(j)}$ denote the mean in the subsample $s_1 = s\setminus \{ j\}$. Following \eqref{Y1}, let
\[
\hat{\theta}_{(j)} = \frac{n-1}{N} \bar{y}_{(j)} + \big( 1 - \frac{n-1}{N} \big) y_j
\]
be the delete-$j$ estimator of $\theta$, where $y_j$ acts as an unbiased estimator of the population mean outside $s_1$. The RB method yields the whole sample mean, denoted by
\[
\hat{\theta}^* = \frac{1}{n} \sum_{j=1}^n \hat{\theta}_{(j)} = \frac{1}{n} \sum_{j=1}^n y_j = \bar{y} 
\] 
Observe that we have $\hat{\theta}^* = \sum_{j=1}^n z_{(j)}/n$, where 
\begin{equation} \label{RB-jack}
z_{(j)} = \frac{1}{N-n} \big( N \hat{\theta}_{(j)} - n \hat{\theta}^* \big) = y_j 
\end{equation}
Thus, the RB estimator $\hat{\theta}^*$ is the mean of an IID sample of observations $z_{(j)}$, for $j = 1, ..., n$, as in the development of classic jackknife variance estimation, so that we obtain 
\[
\widehat{V}(\hat{\theta}^*) = \frac{1}{n(n-1)} \sum_{j=1}^n \big( z_{(j)} - \hat{\theta}^* \big)^2 
\]
Notice that, in this case, the IID observations used for the classic development of jackknife method are given by $z_{(j)} = n\hat{\theta} - (n-1) \hat{\theta}_{(j)} = y_j$ instead of \eqref{RB-jack}, where $\hat{\theta}_{(j)} = \bar{y}_{(j)}$.

\bigskip
For the delete-one RB method based on \eqref{Y1} and \eqref{RB-GR} given auxiliary $\{ x_i : i\in U\}$, we have $\pi_{1i} = \pi_i (n-1)/n$, such that the estimator $b_1$ can be denoted by $b_{(j)}$, based on $s_1 = s\setminus \{ j\}$, where it is simply the delete-$j$ jackknife regression coefficients estimator. Rewrite the corresponding population total estimator $\widehat{Y}_1$ by \eqref{Y1} as
\[
\widehat{Y}_{(j)} = X^{\top} b_{(j)} + \frac{y_j}{\pi_{2j}} - \frac{x_j^{\top} b_{(j)}}{\pi_{2j}} 
\]
such that the RB method yields $\widehat{Y}_{GR}^*$ by \eqref{RB-GR}, as the mean of $\widehat{Y}_{(j)}$ over $j=1, ..., n$. We propose a jackknife variance estimator for $\widehat{Y}_{GR}^*$, given by
\begin{equation} \label{vest-jack}
\widehat{V}(\widehat{Y}_{GR}^*) = \frac{N^2}{n(n-1)} \sum_{j=1}^n \big( z_{(j)} - \frac{1}{n} \sum_{i=1}^n z_{(i)} \big)^2 
\end{equation}
where
\[
z_{(j)} = \frac{1}{N-n} \big( \widehat{Y}_{(j)} - \frac{n}{N} \widehat{Y}_{GR}^* \big) 
\]

Notice that it may be the case under general unequal probability sampling that the conditional inclusion probability $\pi_{2j}$ given $s_1 = s\setminus \{ j\}$ is not exactly known. However, in many situations where the sampling fraction is low, it is reasonable that 
\[
\pi_{2j} \approx \frac{\pi_j}{\sum_{i\not \in s_1} \pi_i} = \frac{\pi_j}{n - \sum_{i\in s_1} \pi_i} \approx \frac{\pi_j}{n ( 1- n_1/N)}
\]
An approximate delete-one RB estimator following \eqref{RB-GR} can then be given as
\begin{equation} \label{approx-RB}
\widetilde{Y}_{GR}^* = X^{\top} \sum_{j=1}^n \frac{b_{(j)}}{n} 
+ \big(1 - \frac{n_1}{N}\big) \sum_{j=1}^n \frac{y_j}{\pi_j} - \big( 1 - \frac{n_1}{N}\big) \sum_{j=1}^n \frac{x_j^{\top}}{\pi_j} b_{(j)}
\end{equation}
with $\widetilde{Y}_{(j)}$ for jackknife variance estimation on replacing $1/\pi_{2j}$ by $n(1-n_1/N)/\pi_j$.  
Meanwhile, the delete-one jackknife replicates of GREG $\widehat{Y}_{GR}$ can be written as
\begin{align*}
& \widehat{Y}_{GR}^{(j)} = X^{\top} b_{(j)} + \frac{n}{n-1} \big( \sum_{i\neq j} \frac{y_i}{\pi_i} - \sum_{i\neq j} \frac{x_i^{\top} b_{(j)}}{\pi_i} \big) \\
& \widehat{Y}_{GR}^{(\cdot)} = \frac{1}{n} \sum_{j=1}^n \widehat{Y}_{GR}^{(j)} = X^{\top} \sum_{j=1}^n \frac{b_{(j)}}{n} + \sum_{i=1}^n \frac{y_i}{\pi_i} 
- \sum_{i=1}^n \frac{x_i^{\top}}{\pi_i} \big( \sum_{j\neq i} \frac{b_{(j)}}{n-1} \big) 
\end{align*}
The estimator $\widehat{Y}_{GR}^{(\cdot)}$ is quite close to the approximate RB-estimator \eqref{approx-RB}; indeed, identical apart from $1-n_1/N$ in the special case of $x_i^{\top}/\pi_i = N/n$. This is not surprising, since the jackknife-based $\widehat{Y}_{GR}^{(\cdot)}$ is an alternative for reducing the bias of the GREG estimator. The difference is that, provided $\pi_{2j}$ is known, the proposed RB method will be exactly design-unbiased, but not the jackknife-based $\widehat{Y}_{GR}^{(\cdot)}$. Finally, the resemblance between $\widetilde{Y}_{GR}^*$ and $\widehat{Y}_{GR}^{(\cdot)}$ is another indication that the relative efficiency of the delete-one RB method is usually not a concern compared to the standard GREG estimator $\widehat{Y}_{GR}$.

\subsection{Monte Carlo RB} 

Exact Rao-Blackwellisation can be computationally expensive, when the cardinality of the subsample space (of $s_1$) is large.  Instead of calculating the RB estimator exactly, consider the Monte Carlo (MC) RB estimator given as follows:
\begin{equation} \label{MC-RB}
\widehat{Y}_{GR}^K = K^{-1} \sum_{k=1}^K \widehat{Y}_{1k} 
\end{equation}
where $\widehat{Y}_{1k}$ is the estimator $\widehat{Y}_1$ based on the $k$th subsample, for $k=1, ..., K$, which are realisations of $s_1$ from $q(s_1 | s)$, such that $\widehat{Y}_{GR}^K$ is a Monte Carlo approximation of $\widehat{Y}_{GR}^*$.

\paragraph{\em Proposition} The estimator $\widehat{Y}_{GR}^K$ is design-unbiased for $Y$, denoted by $E(\widehat{Y}_{GR}^K) =Y$.

\noindent
\emph{Proof:} The result follows from $E(\widehat{Y}_{1k}) =Y$. $\square$

\bigskip \noindent
Adopting a computationally manageable $K$ entails an increase of variance, i.e. $V_q(\widehat{Y}_{GR}^K | s)$, compared to $\widehat{Y}_{GR}^*$, so that the variance of $\widehat{Y}_{GR}^K$ is given by
\begin{equation} \label{V-MC-RB-GR}
V(\widehat{Y}_{GR}^K) = E_1\big( V_2(\widehat{Y}_1 | s_1) \big) - E_p\big( V_q(\widehat{Y}_1 | s) \big) 
+ E_p\big( V_q(\widehat{Y}_{GR}^K | s) \big)
\end{equation}
Due to the IID construction of $\widehat{Y}_{1k}$, an unbiased estimator of $V_q(\widehat{Y}_{GR}^K | s)$ is given by
\[
\widehat{V}_q(\widehat{Y}_{GR}^K | s) = \frac{1}{K(K-1)} \sum_{k=1}^K (\widehat{Y}_{1k} - \widehat{Y}_{GR}^K)^2  
\]
This allows one to control the statistical efficiency of the MC-RB method, i.e. the choice of $K$ is acceptable when $\widehat{V}_q(\widehat{Y}_{GR}^K | s)$ is deemed small enough in practical terms.

\paragraph{\em Proposition} Provided unbiased variance estimator $\widehat{V}(\widehat{Y}_{1k})$ with respect to $p_2(s_2 | s_1)$, i.e. $E_2\big( \widehat{V}(\widehat{Y}_{1k})\big) = V_2(\widehat{Y}_1 | s_1)$, a design-unbiased variance estimator for $\widehat{Y}_{GR}^K$ is given by
\[
\widehat{V}(\widehat{Y}_{GR}^K) = \frac{1}{K} \sum_{k=1}^K \widehat{V}(\widehat{Y}_{1k}) 
- \frac{1}{K} \sum_{k=1}^K (\widehat{Y}_{1k} - \widehat{Y}_{GR}^K)^2  
\]

\noindent
\emph{Proof:} Due to the IID construction of $\widehat{Y}_{1k}$,  $K^{-1} \sum_{k=1}^K \widehat{V}(\widehat{Y}_{1k})$ is an unbiased estimator of the first term on the right-hand side of \eqref{V-MC-RB-GR}, while $(K-1)^{-1} \sum_{k=1}^K (\widehat{Y}_{1k} - \widehat{Y}_{GR}^K)^2$ is an unbiased estimator of the second term. The result follows. $\square$

\bigskip
Finally, for the delete-one RB method, where unbiased variance estimator $\widehat{V}(\widehat{Y}_1)$ is not available now that $|s_2|=1$, a practical option is to first apply the jackknife variance estimator \eqref{vest-jack} to the $K$ samples, as if $\widehat{Y}_{GR}^K$ where the exact RB estimator $\widehat{Y}_{GR}^*$, and then add to it the extra term $\widehat{V}_q(\widehat{Y}_{GR}^K | s)$ for the additional Monte Carlo error. This would allow one to use the Monte Carlo delete-one RB method in general.

\section{Unbiased non-linear learning} \label{non-linear}

In this section we consider design-unbiased estimation in survey sampling, which builds on arbitrary ML technique that can be non-linear as well as non-parametric.

\subsection{Design-unbiased ML for survey sampling}

Denote by $M$ the model or algorithm that aims to predict $y_i$ given $x_i$. Let $s_1$ be the training set, and $s_2 = s\setminus s_1$ the test set. Let $\widehat{M}$ be the trained model based on $\{ (x_i, y_i) : i\in s_1\}$, yielding $\mu(x_i, s_1)$ as the corresponding \emph{$M$-predictor} of $y_i$ given $x_i$. Apply the trained model to $i\in s_2$ yields the \emph{prediction errors} of $\widehat{M}$ \emph{conditional on $s_1$}, denoted by $e_i = y_i - \mu(x_i, s_1)$.
In contrast, the \emph{same} discrepancy is referred to as the \emph{residuals} of $\widehat{M}$, when it is calculated for $i\in s_1$, denoted by $\hat{e}_i = y_i - \mu(x_i, s_1)$, including when the training set $s_1$ is equal to $s$. In standard ML, the errors in the test set are used to select different trained algorithms, or to assess how well a trained algorithm can be expected to perform when applied to the units with unknown $y_i$'s. 

From an inference point of view, a basic problem with the standard ML approach above arises because one needs to be able to `extrapolate' the information in $\{ e_i : i\in s_2\}$ to the units outside $s$, in order for supervised learning to have any value at all. This is simply because $\{ y_i : i\in s\}$ are all observed and prediction in any form is \emph{unnecessary} for $i\in s$. No matter how the training-test split is carried out, one cannot ensure valid $\mu(x_k, s_1)$ for $k\not\in s$, \emph{unless} $s$ is selected from the entire reference set of units, i.e. the population $U$, in some non-informative (or representative) manner. This is the well-known problem of observational studies in statistical science, which is sometimes recast as the problem of concept drift in the ML literature (e.g. Tsymbal, 2004).

A \emph{design $pq$-unbiased} approach to \emph{M-assisted estimation} of population total $Y$ can be achieved with respective to \begin{itemize}
\item[(i)] a probability sample $s$ from $U$, with probability $p(s)$, and
\item[(ii)] a probabilistic scheme $q(s_1 |s)$ for the training-test split $(s_1, s_2)$ given $s$. 
\end{itemize}
Explicitly, let $\widehat{Y}_{1M}$ be the estimator of $Y$ obtained from the realised sample $s$ and subsample $s_1$ given the model $M$. It is said to be design $pq$-unbiased for $Y$, provided
\[
E_{pq}(\widehat{Y}_{1M}) = \sum_s p(s) \sum_{s_1 \subset s} q(s_1 |s) \widehat{Y}_{1M} = Y
\]
where $E_{pq}$ is the expectation of $\widehat{Y}_{1M}$ over all possible $(s, s_1)$. Replacing the linear predictor $x_i^{\top} b_1$ in \eqref{Y1} by any $M$-predictor $\mu(x_i, s_1)$ trained on $s_1$, we obtain
\begin{equation} \label{Y1M}
\widehat{Y}_{1M} = \sum_{i\in s_1} y_i + \sum_{i\in U\setminus s_1} \mu(x_i, s_1) + \sum_{i\in s_2} e_i/\pi_{2i} 
\end{equation}

\paragraph{\em Proposition} $\widehat{Y}_{1M}$ by \eqref{Y1M} is design $pq$-unbiased for $Y$ using an arbitrary model $M$.

\bigskip
The proof is parallel to that for $\widehat{Y}_1$ by \eqref{Y1}, only that $\mu(x, s_1)$ is now based on any chosen model $M$. It is important to point out that the purpose here is to estimate $Y$ at the population level, instead of individual prediction \emph{per se}. Indeed, $\widehat{Y}_{1M}$ is design-unbiased, regardless $M$ is a strong or weak learner. The underlying probabilistic mechanism consists of two necessary elements: $p(s)$ ensures valid extrapolation of learning to the units outside $s$, since otherwise completely model-based prediction $\sum_{i\in s} y_i + \sum_{i\in U\setminus s} \mu(x_i, s)$ has no guaranteed relevance to $Y$ no matter how the training set $s_1$ is chosen or how $M$ is selected, whereas subsampling $q(s_1 |s)$ is required to be able to project the errors in $s_2$ to the aggregated level, since projecting the residuals in $s$ in the manner of GREG estimator $\widehat{Y}_{GR}$ (i.e. without the training-test split) would not achieve unbiasedness exactly.

\subsection{Subsampling RB and bootstrap aggregating}

There is a natural affinity between the subsampling RB method and bootstrap aggregating (i.e. \emph{bagging}). Bagging is originally devised to improve unstable leaners (Breiman, 1996a; 1996b) for individual prediction, where the aggregation averages the learner over bootstrap replicates of the training set. The argument can be adapted to design-based population-level estimation. Let $\widehat{Y}_M = \psi(s;M)$ be an $M$-assisted estimator of $Y$, which varies over different samples $s$. Insofar as $\{ y_i : i\in U \}$ are treated as unknown constants and $\widehat{Y}_M$ is uniquely determined given $\{ (y_i, x_i) : i\in s\}$, the only variation of $\widehat{Y}_M$ derives from that of the sample $s$. For some model $M$, such as regression tree with random feature selection, there exists an extra variation of $\widehat{Y}_M$ given $s$. In any case, let the expectation of $\widehat{Y}_M$ be
\[
\psi_M = E(\widehat{Y}_M) = E\big( \psi(s;M) \big) 
\]
over all possible $s$ and additional randomness given $s$. We have
\[
E\big( (\widehat{Y}_M - Y)^2 \big) = (\psi_M - Y)^2 + E\big( (\widehat{Y}_M -\psi_M)^2 \big)
\]
since $E(\widehat{Y}_M -\psi_M) =0$ by definition. Thus, $\psi_M$ has always a smaller mean squared error than $\widehat{Y}_M$. Notice that in reality bagging is ``caught in two currents'' (Breiman, 1994): the improvement can be appreciable if $\widehat{Y}_M$ is unstable, whereas the additional estimation of $\psi_M$ by bagging may not be worthwhile if $\widehat{Y}_M$ is a stable learner to start with.

It is clear from the above that, while it can reduce the variance of unbagged predictor, bagging does not affect the potential bias, now that it aims at replacing $\widehat{Y}_M = \psi(s; M)$ by its expectation $\psi_M$. The subsampling RB method is more effectual than bagging in the following sense: on the one hand, it leads generally to design-unbiased estimation of $Y$, which does not result from bagging alone; on the other hand, Rao-Blackwellising $\widehat{Y}_{1M}$ reduces its variance, even when it is based on a stable learner, such as $\mu(x_i, s_1) = x_i^{\top} b_1$, which bagging does only for unstable $\widehat{Y}_M$. Replacing $\widehat{Y}_1$ in \eqref{RB-GR} with $\widehat{Y}_{1M}$ given by \eqref{Y1M} generally, the subsampling RB $M$-assisted estimator of $Y$ is given by
\begin{equation} \label{RB-M}
\widehat{Y}_M^* = E_q(\widehat{Y}_{1M} | s)
\end{equation}

\paragraph{\em Proposition} The subsampling RB $M$-assisted estimator $\widehat{Y}_M^*$ by \eqref{RB-M} is design $p$-unbiased, or simply design-unbiased, for $Y$ using an arbitrary model $M$.

\bigskip
The proof is exactly parallel to that for $\widehat{Y}_{GR}^*$ by \eqref{RB-GR}. Notice that Rao-Blackwellisation of $\widehat{Y}_{1M}$ with respect to $q(s_1 |s)$ can accommodate straightforwardly any additional variation given $s_1$ due to the chosen model $M$. For example, given a subsample $s_1$, one can grow a regression tree with random feature selection. Despite the resulting $\mu(x, s_1)$ is not fixed for the given $s_1$, the corresponding $\widehat{Y}_{1M}$ is still design $pq$-unbiased, because it is conditionally unbiased for $Y$ given $s_1$ \emph{and} the outcome of random feature selection, and $Y$ is a constant with respect to subsampling of $s_1$ and random feature selection given $s_1$. 

Finally, Monte Carlo subsampling RB is operationally similar to bagging, involving about the same amount of computation effort. In bagging, one draws a bootstrap replicate sample from $s$; whereas in subsampling RB, one resamples $s_1$ from $s$ according to $q(s_1 |s)$. In either case, one trains the model based on the resample. Repeating the two steps $K$ times yields the bagged predictor by bagging, and the MC-RB estimator by subsampling RB. The choice of $K$ balances between numerical and statistical efficiency.

\subsection{Design consistency}

Provided $n_2 = |s_2| \geq 2$, let $\widehat{V}(\widehat{Y}_{1M,k})$ be an unbiased variance estimator with respect to $p_2(s_2 | s_1)$, i.e. $E_2\big( \widehat{V}(\widehat{Y}_{1M,k} )\big) = V_2(\widehat{Y}_{1M,k})$, for $K$ subsamples $k=1, ..., K$. A design-unbiased variance estimator for MC-RB estimator $\widehat{Y}_M^K$ is given by
\begin{equation} \label{var-MC-RB-M}
\widehat{V}(\widehat{Y}_M^K) = \frac{1}{K} \sum_{k=1}^K \widehat{V}(\widehat{Y}_{1M,k}) 
- \frac{1}{K} \sum_{k=1}^K (\widehat{Y}_{1M,k} - \widehat{Y}_M^K)^2  
\end{equation}
similarly as for $\widehat{Y}_{GR}^K$. It is an open question at this stage how to determine the efficient subsampling scheme $q(s_1 |s)$, including the choice $n_1$. Although given the simplicity and practical advantage of the delete-one GREG-assisted $\widehat{Y}_{GR}^K$, any other $M$-assisted estimator would not be worth considering, \emph{unless} it has clearly a smaller estimated variance. 

A design-unbiased $M$-assisted estimator is consistent, provided its sampling variance tends to 0 asymptotically, as $n, N \rightarrow \infty$. Since this is the case with delete-one GREG-assisted $\widehat{\bar{Y}}_{GR}^*$ of population mean $\bar{Y} = Y/N$, and that in practice one would only admit any alternative estimator that has an even smaller variance, design consistency is not a worrisome issue for design-unbiased $M$-assisted estimation in applications.

Meanwhile, we cannot find any direct references in the literature, concerning the design consistency of ML techniques. For example, Gordon and Olshen (1978, 1980) establish consistency of recursive partitioning algorithms, such as regression tree, provided IID training set. Toth and Eltinge (2011) extend their result, allowing sampling design \emph{in addition} to the IID super-population model $M$, such that the consistency of regression tree for individual prediction, based on samples selected from $p(s)$, is not purely design-based, but requires the super-population model to hold in addition.

In the standard ML literature, asymptotic results are typically derived under stability conditions. Bousquet and Elisseeff (2002) establish uniform stability condition for Regularisation algorithms. Mukherjee et al. (2006) pay special attention to empirical risk minimisation algorithms. Both these works are directed at \emph{individual-level} predictor from IID training set, denoted by $\mathcal{S} = \{ (x_i, y_i) : i =1, ..., n\}$, asymptotically as $n\rightarrow \infty$. Let $\mathcal{Z} = (x,y)$ be generically the random variables from the relevant distribution. Let $\mu(x, \mathcal{S})$ be a given predictor trained on $\mathcal{S}$. Its prediction mean squared error is $E_{\mathcal{Z}}\big[ (y - \mu(x, \mathcal{S})^2 \big]$. Expectation with respect to $\mathcal{S}$ is needed in addition for the stability definitions. 

Different definitions of stability are needed under the $pq$-design-based approach to population-level estimation, where $\{ (x_i, y_i) : i\in U\}$ are treated as constants and only the sample $s$ is random. Below we consider first the delete-one RB estimator \eqref{RB-M} under the special case of SRS and, then, under general unequal probability sampling design.

\subsubsection{Stability condition: SRS}

Let $s_j = s \setminus \{ j\}$ be the delete-$j$ sample. Let $s_{ij} = s\setminus \{i,j\}$ be the delete-$ij$ sample. Let $\mu(x, s_j)$ be the $M$-predictor given $x$, which is trained on $s_j$, and $\mu(x, s_{ij})$ that on $s_{ij}$. We define $\mu(x, s)$ to be \emph{twice q-stable}, if 
\begin{equation}\label{twice-q}
\mu(x_k, s_j) - \mu(x_k, s) ~\stackrel{P}{\rightarrow}~ 0 \quad\text{and}\quad
\mu(x_k, s_{ij}) - \mu(x_k, s_j) ~\stackrel{P}{\rightarrow}~ 0 
\end{equation}
i.e. convergence in probability, as $n, N\rightarrow \infty$ asymptotically, for any $i,j\in s$ and $k\in U$, where $s_j$ results from delete-one $q$-sampling from $s$, and $s_{ij}$ from recursive $q$-sampling where one randomly deletes $i\in s_j$. Notice that the first part of \eqref{twice-q} is analogous to the `point-wise hypothesis leave-one-out stability' of Mukherjee et al. (2006).

\paragraph{\emph{Theorem 1:}} The delete-one RB estimator \eqref{RB-M} is consistent for population mean $\bar{Y}$ under SRS, as $n, N\rightarrow \infty$, given twice $q$-stability and $y_i - \mu(x_i, s) = O(1)$ for any $s$.

\noindent
\emph{Proof:} We have $V_p(\widehat{Y}_M^*) = E_1\big( V_2(\widehat{Y}_{1M} | s_1)\big) - E_p\big( V_q(\widehat{Y}_{1M} |s) \big)$ as by \eqref{V-RB-GR}, where
\begin{align*}
\widehat{Y}_{1M}(s_j) & = \sum_{k\in s_j} y_k + \sum_{k\not \in s_j} \mu(x_k, s_j) + (N-n+1) z_j(s_j) \\
& = \sum_{k \in s} y_k + \sum_{k \not \in s} \mu(x_k, s_j) + (N - n) z_j(s_j) 
\end{align*}
and $z_k(s_j) = y_k - \mu(x_k, s_j)$ for any $k$ and delete-$j$ sample $s_j$. Under SRS, we have
\[
V_2\big( \widehat{Y}_{1M}(s_j) | s_j\big) = (N-n+1) \sum_{k\in s_j^c} \big( z_k(s_j) - \bar{Z}_{s_j^c}(s_j) \big)^2 
\]
where $\bar{Z}_{s_j^c}(s_j) = \sum_{k\in s_j^c} z_k(s_j)/(N-n)$ and $s_j^c = U\setminus s_j$. By \eqref{twice-q}, we have
\[
z_k(s_j) - \bar{Z}_{s_j^c}(s_j) = [1+ o_p(1)] \big( z_k(s_{ij}) - \bar{Z}_{s_j^c}(s_{ij}) \big)
\]
for any $i\in s_j$, where $\bar{Z}_{s_j^c}(s_{ij}) = \sum_{k\in s_j^c} z_k(s_{ij})/(N-n)$, and, averaged over all $i\in s_j$,  
\begin{align*}
V_2\big( \widehat{Y}_{1M}(s_j) | s_j\big) & = [1+o_p(1)] (N-n+1) (N-n) \{ \sum_{i\in s_j} \frac{1}{n-1} \widehat{V}_{s_j^c} \} \\
\widehat{V}_{s_j^c} & = \frac{1}{n_{s_j^c} -1} \sum_{k\in s_j^c} \big( z_k(s_{ij}) - \bar{Z}_{s_j^c}(s_{ij}) \big)^2 
\end{align*}
where $n_{s_j^c} = |s_j^c| = N-n+1$. One can consider $\widehat{V}_{s_j^c}$ as an unbiased estimator of 
\[
\tau(s_{ij}) = \frac{1}{N_{s_{ij}^c} -1} \sum_{k\in s_{ij}^c} \big( z_k(s_{ij}) - \bar{Z}_{s_{ij}^c}(s_{ij}) \big)^2
\] 
where $N_{s_{ij}^c} = |s_{ij}^c| =N-n+2$, i.e. the population variance of $z_k(s_{ij})$ in $s_{ij}^c$, based on SRS sample $s_j^c$ from $s_{ij}^c$ conditional on $s_{ij}$, since $s_{ij}^c = s_j^c \cup \{ i\} = U\setminus s_{ij}$, such that 
\[
E_1\big( \sum_{i\in s_j} \frac{1}{n-1} \widehat{V}_{s_j^c} \big) = E_{s_{ij}}\big( E_{i}(\widehat{V}_{s_j^c}  | s_{ij}) \big) 
= E_{s_{ij}}\big(\tau(s_{ij})\big) ~.
\]
Given $y_k - \mu(x_k, s) = O(1)$ for any $s$ and $k\in U$, we obtain 
\[
E_1\big( V_2(\widehat{Y}_{1M} | s_1) \big) = [1+ o(1)] (N-n+1) (N-n) E_{s_{ij}} \big(\tau(s_{ij}) \big) ~.
\]
Next, for $V_q(\widehat{Y}_{1M} |s)$, we notice that, by $\mu(x_k, s_j) - \mu(x_k, s) ~\stackrel{P}{\rightarrow}~ 0$ in \eqref{twice-q}, 
\[
\widehat{Y}_{1M}(s_j) = [1 + o_p(1)] \big[ \sum_{k \in s} y_k + \sum_{k \not \in s} \mu(x_k, s) + (N - n) z_j(s_j) \big]
\]
where $V_q\big( z_j(s_j) |s\big) = \sum_{i\in s} \big( z_i(s_i) - \bar{z}(s) \big)^2/n$, for $\bar{z}(s) = \sum_{j\in s} z_j(s_j)/n$.
In Sen-Yates-Grundy type expression using pairwise differences, we can write
\begin{align*}
V_q\big( z_j(s_j) |s\big) & = \frac{n-1}{n} \{ \big( \frac{n(n-1)}{2} \big)^{-1} \sum_{i< j\in s} \frac{1}{2} \big( z_i(s_i) - z_j(s_j) \big)^2 \} \\
& = \frac{n-1}{n} \{ \big( \frac{n(n-1)}{2} \big)^{-1} \sum_{i< j\in s} \widehat{V}_{ij} \} [1 + o_p(1)]
\end{align*}
where $\widehat{V}_{ij} = \frac{1}{2} \big( z_i(s_{ij}) - z_j(s_{ij}) \big)^2$, by $\mu(x_k, s_{ij}) - \mu(x_k, s_j) ~\stackrel{P}{\rightarrow}~ 0$ in \eqref{twice-q}. One can consider $\widehat{V}_{ij}$ as an unbiased estimator of $\tau(s_{ij})$, i.e. the population variance of $z_k(s_{ij})$ in $s_{ij}^c$, based on SRS sample $s_2' = \{ i, j\}$ from $s_{ij}^c$ conditional on $s_1' = s_{ij}$. Moreover, one can view the expression in last brackets $\{ \}$ above as $E_{q'}(\widehat{V}_{ij} | s)$ with respect to $q'(s_{ij} |s)$, such that 
\[
E_p\big( \{ \cdot \} \big) = E_p\big( E_{q'}(\widehat{V}_{ij} | s) \big) = E_{s_1'} \big( E_{s_2'}( \widehat{V}_{ij} |s_{ij}) \big) 
= E_{s_1'} \big( \tau(s_{ij}) \big) \equiv E_{s_{ij}} \big( \tau(s_{ij}) \big) ~.
\]
Given $y_k - \mu(x_k, s) = O(1)$ for any $s$ and $k\in U$, we obtain
\[
E_p\big(V_q(\widehat{Y}_{1M} |s) \big) = [1 + o(1)] (N-n)^2 \frac{n-1}{n} E_{s_{ij}} \big( \tau(s_{ij}) \big)~.
\]
Finally, the result follows from $E_1\big( V_2(\widehat{Y}_{1M} |s_1)\big)$ and $E_p\big( V_q(\widehat{Y}_{1M} |s) \big)$ above, since 
\[
V_p(\widehat{\bar{Y}}_M^*) = (1-\frac{n}{N}) \frac{1}{n} E_{s_{ij}}\big(\tau(s_{ij})\big) + o(1) ~.\quad \square
\]

\subsubsection{Stability condition: Unequal probability sampling}

For general unequal probability sampling, we define the following the stability conditions. First, we define $\mu(x, s)$ to be simply \emph{q-stable} if, for any $j\in s$ and $k\in U$, we have 
\begin{equation}\label{q-stable}
\mu(x_k, s_j) - \mu(x_k, s) ~\stackrel{P}{\rightarrow}~ 0 
\end{equation}
asymptotically as $n, N\rightarrow \infty$, where $s_j$ results from delete-one subsampling $q(s_j | s)$. Next, we define $\mu(x ,s)$ to be \emph{p-stable} for the delete-one RB method, if 
\begin{equation}\label{p-stable}
\frac{1}{n} \sum_{j\in s} \frac{\widehat{N}_j}{N} \mu(x_j, s) - \frac{1}{N} \sum_{k\in U} \mu(x_k, s) ~\stackrel{P}{\rightarrow}~ 0
\end{equation}
where $\widehat{N}_j = \pi_{2j}^{-1} + (n-1)$ is an estimator of $N$ based on $s_2 = \{j\}$. Notice that, given $q$-stability \eqref{q-stable}, it is possible to replace $p$-stability \eqref{p-stable} by a $pq$-stability condition
\[
\frac{1}{n} \sum_{j\in s} \frac{\widehat{N}_j}{N} \mu(x_j, s_j) - \frac{1}{N} \sum_{k\in U} \mu(x_k, s_j) ~\stackrel{P}{\rightarrow}~ 0 
\]
which reduces to $\sum_{j\in s} \mu(x_j, s_j)/n - \sum_{k\in U} \mu(x_k, s_j)/N \stackrel{P}{\rightarrow} 0$ under SRS, and resembles the IID `expected-leave-one-out stability' of Mukherjee et al. (2006): the first term above is the empirical average in the observed set in both definitions, whereas for the second term here we replace averaging over $\mathcal{Z}$ in the IID setting by that over the population distribution function, which places point mass $1/N$ on each $k\in U$. 

Some regularity condition on the sampling design $p(s)$ is needed to for the general situation. Let $\widehat{\bar{Y}}_j = y_j \widehat{N}_j/N$ be the \emph{leave-one-out (LOO) HT} estimator of population mean $\bar{Y}$ based on $s_2 = \{j\}$. We define the sampling design to be \emph{LOO-consistent}, if 
\begin{equation} \label{q-consistent} 
\frac{1}{n} \sum_{j\in s} \widehat{\bar{Y}}_j ~\stackrel{P}{\rightarrow}~ \bar{Y}
\end{equation}
asymptotically as $n, N\rightarrow \infty$. The condition is specified for the LOO-RB-HT estimator, where $\sum_{j\in s} \widehat{\bar{Y}}_j /n = E_q(\widehat{\bar{Y}}_j |s)$. Under SRS, $\sum_{j\in s} \widehat{\bar{Y}}_j/n =\sum_{j\in s} y_j/n$ is the sample mean, which converges to $\bar{Y}$ in probability, provided $y_i = O(1)$ for all $i\in U$. We emphasise that the condition \eqref{q-consistent} concerns only the sampling design $p(s)$, since it is formulated in terms of the $y$-values alone, i.e. based on an `empty' $M$-predictor,  so to speak.

\paragraph{\emph{Theorem 2:}} The delete-one RB estimator \eqref{RB-M} is consistent for population mean $\bar{Y}$, as $n, N\rightarrow \infty$, provided $q$- and $p$-stabilities, and LOO-consistent sampling design $p(s)$.

\noindent
\emph{Proof:} Given the delete-$j$ sample $s_j$ under any general sampling design, we can write
\begin{align*}
\widehat{Y}_{1M}(s_j) & = \sum_{i\in s} y_i + \sum_{k \notin s} \mu(x_k, s_j) + (\pi_{2j}^{-1} -1) \big( y_j - \mu(x_j, s_j) \big) \\
& = \sum_{i\in s} \big( y_i - \mu(x_i, s_j) \big) + \sum_{k\in U} \mu(x_k, s_j) + (\pi_{2j}^{-1} -1) \big( y_j - \mu(x_j, s_j) \big) 
\end{align*}
where $\pi_{2j}$ is the conditional probability of selecting $j$ from $s_j^c$ given $s_j$. Given $q$-stability \eqref{q-stable}, i.e. $\mu(x, s_j) = \mu(x, s) + o_p(1)$, the RB-estimator of the population mean is 
\begin{align*}
\widehat{\bar{Y}}_M^* & = \frac{1}{N} \sum_{k\in U} \mu(x_k, s) 
+ \frac{1}{n} \sum_{j\in s} \frac{\widehat{N}_j}{N} \big( y_j - \mu(x_j, s) \big) + o_p(1)\\
& = \{ \frac{1}{N} \sum_{k\in U} \mu(x_k, s) - \frac{1}{n} \sum_{j\in s} \frac{\widehat{N}_j}{N} \mu(x_j, s) \} 
+ \big[ \frac{1}{n} \sum_{j\in s} \widehat{\bar{Y}}_j \big]+ o_p(1) ~.
\end{align*}
The result follows from applying the $p$-stability condition \eqref{p-stable} to the expression in the brackets $\{ \}$, and the LOO-consistency condition \eqref{q-consistent} to that in $[~ ]$. $\square$

\section{Simulations} \label{simulation}

Below we present and discuss some simulation results of the delete-one RB (or LOO-RB) method, and the associated jackknife variance estimator described in Section \ref{LOO-RB-GREG}. The HT and some GREG estimators are computed for comparisons. The target is always the population mean (denoted by $\theta$) in a given set-up. The simulations proceed as follows.
\begin{itemize}[leftmargin=6mm]
\item[-] $B$ samples (usually $B=100$) are drawn independently from the given fixed population according to a specified sampling design. 
\item[-] We obtain an estimate $\hat{\theta}_{(b)}$ based on each sample, for $b=1, ..., B$. In particular, for the LOO-RB method, we calculate its associated jackknife variance estimate $v_{(b)}$.  
\item[-] An estimate of $E(\hat{\theta})$ over repeated sampling is $\hat{\bar{\theta}} = \sum_{b=1}^B \hat{\theta}_{(b)}/B$, with associated Monte Carlo error $\sqrt{v/B}$, where $v=\sum_{b=1}^B (\hat{\theta}_{(b)} - \hat{\bar{\theta}})^2/(B-1)$. An estimate of its bias is $\hat{\bar{\theta}} - \theta$; an estimate of its root mean squared error (RMSE) is $\{ \sum_{b=1}^B (\hat{\theta}_{(b)} - \theta)^2/B\}^{1/2}$.
\item[-] Similarly for the bias and RMSE of the variance estimator $v_{(b)}$, except that the true variance of the LOO-RB method is unknown and is replaced by its estimate $v$. 
\end{itemize}
Now that the HT estimator and the LOO-RB methods are unbiased, an inspection of their respective simulation-based bias estimates and the associated Monte Carlo errors can usually provide adequate information, in order to judge whether a certain conclusion of the results is warranted given the actual number of simulations.

\subsection{Simulations with synthetic data}

The GREG estimator has become the standard-bearer in practical survey sampling in the past three decades. Using simple simulations below, we would like to gain some basic appreciation of the pros and cons of the corresponding LOO-RB-GREG estimator, given by \eqref{RB-GR}, under the proposed unbiased learning approach. Small synthetic populations are generated based on only two regressors. The first regressor $x_1$ follows a log-normal distribution with mean and variance both set to one. The second regressor $x_2$ follows a Poisson distribution with mean $5$. The target $y$-variable in each setting is generated as the absolute value of a certain function of $x_1$ and $x_2$ plus a regression error.

\begin{table}[htbp]
\centering
\caption{Simulation results of HT, GREG and LOO-RB-GREG estimator, by two different sampling designs. Monte Carlo errors of bias estimates in parentheses.}
\begin{tabular}{| l | r | r | r | r |} \hline
& \multicolumn{2}{|c|}{Simple Random Sampling} & \multicolumn{2}{c|}{Probability Proportional to $x_2$} \\ \cline{2-5}
Estimator & Bias (MC Error) & RMSE & Bias (MC Error) & RMSE \\ \hline
HT  & 0.08 \hspace{5mm} (0.19) & 1.91 &  0.12 \hspace{5mm} (0.20) & 2.02 \\ \hline
GREG & -0.09 \hspace{5mm} (0.13) & 1.29  & 0.03 \hspace{5mm} (0.16) & 1.60\\ \hline
LOO-RB-GREG & 0.10 \hspace{5mm} (0.14) & 1.39 & 0.16 \hspace{5mm} (0.16) & 1.63\\ \hline
Variance by jackknife & 0.52 \hspace{5mm} (0.10) & 1.09 & 4.73 \hspace{5mm} (1.02) & 11.23\\ \hline
\end{tabular} \label{tab:simple}
\end{table}

We start with a setting where the GREG estimator should have a negligible or very small bias. Let the population size be 200, and let the target survey $y$-variable be the absolute value of $1.5 x_1 + x_2 + \epsilon$, where $\epsilon$ follows a normal distribution with zero mean and variance that is a quarter of the variance of $x_1$. Let the sample size be 20. Two sampling designs are used: SRS, or conditional Poisson sampling with probabilities proportional to $x_2$ as the size variable. The results are given in Table \ref{tab:simple}.

It can be seen that under both sampling designs, GREG and LOO-RB-GREG have essentially the same efficiency, and both outperform HT estimation. Recall that the bias of the GREG estimator is negligible in this scenario because of the underlying linear population model. Clearly, the jackknife variance estimator \eqref{vest-jack}, which is derived as a direct analogy to the IID-sample situation, needs to be modified for unequal probability sampling designs such as the conditional Poisson sampling here.

\begin{table}[htbp]
\centering
\caption{Simulation results of HT, GREG and LOO-RB-GREG estimator under SRS, but two different population models. Monte Carlo errors of bias estimates in parentheses.}
\begin{tabular}{| l | r | r | r | r |} \hline
& \multicolumn{2}{|c|}{$V(y) \propto x_1$, $n=5$} & \multicolumn{2}{c|}{Non-linear, $V(y) \propto \sqrt{x_1}$, $n=20$} \\ \cline{2-5}
Estimator & Bias (MC Error) & RMSE & Bias (MC Error) & RMSE \\ \hline
HT  & -0.46 \hspace{5mm} (0.50) & 5.03      & 0.95 \hspace{5mm} (1.26) & 12.57 \\ \hline
GREG & -0.82 \hspace{5mm} (0.41) & 4.16   & -2.41 \hspace{5mm} (0.51) & 5.62\\ \hline
LOO-RB-GREG & -0.68 \hspace{5mm} (0.77)  & 7.69   & 0.68 \hspace{5mm} (0.86) & 8.62\\ \hline
Variance by jackknife & -13.22 \hspace{3mm} (15.75) & 157.31  & -7.36 \hspace{3mm} (10.50) & 104.76\\ \hline
\end{tabular} \label{tab:problem}
\end{table}

Consider now two potentially problematic settings. First, we introduce heteroscedasticity by make the variance of the $y$-variable  proportional to $x_1$, while reducing the sample size at the same time, where $n=5$ (from $N = 100$). The results are given in the left part of Table \ref{tab:problem}. The LOO-RB-GREG is the least efficient estimator here: the heteroscedasticity setting increases the variance of $\widehat{Y}_1$ based on each subsample, whereas the small sample size implies RB averaging over only 5 subsample estimates (instead of 20 above). The RMSE of the jackknife variance estimator is much bigger for similar reasons. 

Next, reverting to $(N, n) = (200, 20)$, we generate the target $y$-variable non-linearly as the absolute value of $0.5 x_1 + 0.25 x_1^2 + x_2 + \epsilon$, where $\epsilon$ follows a normal distribution with zero mean and variance proportional to $\sqrt{x_1}$. The results under SRS are given in the right part of Table \ref{tab:problem}. The GREG estimator has now a relatively large bias, which is removed by the LOO-RB-GREG estimator. However, the unbiased learning estimator loses efficiency compare to GREG in terms of the MSE, although it is still much better than the HT estimator. The performance of the variance estimator is similar as before. 

These results illustrate the basic pros and cons of delete-one RB-GREG vs. standard GREG estimation. On the one hand, the GREG estimator may suffer from non-negligible bias, e.g. because one applies the assisting linear model in a routine manner without conducting careful model diagnostics as one should, whereas the unbiased learning approach avoids the bias by definition. On the other hand, delete-one subsampling may suffer from loss of efficiency given heteroscedastic observations in very small samples.

\subsection{Simulations with real data}

The population consists of a sample of about 17000 small and medium-sized enterprises from the Spanish Structural Business Survey (SSBS). As the target variables we consider three survey variables collected in the SSBS: Turnover, Total personnel expenses and Total procurements of goods and services. Seventeen variables from the administrative corporate income tax data are imported as the regressors. One of them is turnover, although for many enterprises the turnover from tax data will be different to the turnover from SSBS by definition; in addtion the two observed values may differ because of registration delays or other operational reasons.  The estimators to be considered are: HT, GREG1 with one regressor (turnover), GREG17 with the seventeen regressors (as main effects), LOO-RB-GREG1 with one regressor (turnover), and LOO-RB random forest (RF) with seventeen features. When only one regressor is used, RF is not a good option to be included here. Jackknife variance estimation is applied to the two SRB estimators.

\subsubsection{Turnover}

This case is interesting because turnover (from tax data) is one of the regressors. We consider SRS and stratified SRS designs. For the latter, three strata are created by the number of employees, which is a commonly used stratification variable in SBS, although the actual designs always have many other complicating details in practice. The stratum sample sizes are allocated proportionally to the stratum population sizes. The total sample size is 10\% of the population under both the designs. The simulation results are given in Table \ref{tab:turn}, similarly as before and suitably scaled for presentation.

\begin{table}[htbp]
\centering
\caption{Simulation results for Turnover estimation by different estimators under two sampling designs. Monte Carlo errors of bias estimates in parentheses.}
\begin{tabular}{| l | r | r | r | r|} \hline
& \multicolumn{2}{|c|}{SRS} & \multicolumn{2}{c|}{Stratified SRS} \\ \cline{2-5}
Estimator & Bias (MC Error) & RMSE & Bias (MC Error) & RMSE \\ \hline
HT  & -0.22  \hspace{5mm} (0.47) & 4.65          & -0.36  \hspace{5mm} (0.32)  & 3.20\\ \hline
GREG1  &  0.01 \hspace{5mm} (0.26) & 2.56   & 0.30 \hspace{5mm} (0.22)  &  2.16\\ \hline
LOO-RB-GREG1  & -0.21 \hspace{5mm} (0.27) & 2.68 & 0.05 \hspace{5mm} (0.23) & 2.31 \\ \hline
Variance LOO-RB-GREG1 & -1.02 \hspace{5mm} (0.32) & 3.37  & 0.60 \hspace{5mm} (0.32) & 3.27 \\ \hline
GREG17 & 0.42 \hspace{5mm} (0.22) & 2.30   & 0.79 \hspace{5mm} (0.75)  &  7.47\\ \hline
LOO-RB-RF & -0.19 \hspace{5mm} (0.14) & 1.37 & -0.10 \hspace{5mm} (0.14) & 1.38\\ \hline
Variance LOO-RB-RF & 0.09 \hspace{5mm} (0.03) & 0.34 & -0.04 \hspace{5mm} (0.04) & 0.38 \\ \hline
\end{tabular} \label{tab:turn}
\end{table}

The LOO-RB-RF (by random forest) is the most efficient estimator under SRS. It is more efficient than the GREG17 estimator, because RF yields a better prediction model than simple linear regression using all the regressors as main effects. In fact, the GREG17 estimator introduces a small bias compared to the GREG1 estimator, and is only more efficient by a small margin. The LOO-RB-GREG1 estimator has about the same efficiency as the GREG1 estimator. Compared to the simple simulation results earlier, heteroscedastic variance does not cause loss of efficiency to the LOO-RB method here, because the sample size is large enough. The jackknife variance estimator has no statistically significant bias for the LOO-RB-RF estimator, but it has a small negative bias for the LOO-RB-GREG1 estimator.  

Next, under the stratified SRS design, the LOO-RB-RF is again the most efficient estimator. Its RMSE is about the same as under SRS, which is not surprising given proportional allocation of stratum sample sizes, because RF is able to account for the design size variable using the auxiliary information in the 17 regressors. In contrast, the GREG17 estimator actually loses efficiency and does not behave well here, which again illustrates that applying the GREG estimator without appropriate attention to model diagnostics can be counter-productive in practice. The relative performance of the simple GREG1 estimator and its unbiased counterpart LOO-RG-GREG1 is similar as under SRS, and both are slightly more efficient under the stratified design. The jackknife variance estimators perform similarly as under the SRS design.

\subsubsection{Other target variables}

Simulation results for the other two target variables under the stratified SRS design are given in Table \ref{tab:other}. 

\begin{table}[htbp]
\centering
\caption{Simulation results for Total personal expenses and Total procurements under stratified SRS design. Monte Carlo errors of bias estimates in parentheses.}
\begin{tabular}{| l | r | r | r | r|} \hline
& \multicolumn{2}{|c|}{Total personal expenses} & \multicolumn{2}{c|}{Total procurements} \\ \cline{2-5}
Estimator & Bias (MC Error) & RMSE & Bias (MC Error) & RMSE \\ \hline
HT  & 0.03  \hspace{5mm} (0.04) & 0.58          & -0.05  \hspace{5mm} (0.29)  & 2.88\\ \hline
GREG1  &  0.09 \hspace{5mm} (0.04) & 0.60   & -0.07 \hspace{5mm} (0.21)  &  2.08\\ \hline
LOO-RB-GREG1  & 0.06 \hspace{5mm} (0.04) & 0.60    & -0.20 \hspace{5mm} (0.21) & 2.14 \\ \hline
Variance LOO-RB-GREG1 & 0.02 \hspace{5mm} (0.01) & 0.03  & 0.03 \hspace{5mm} (0.14) & 1.37 \\ \hline
GREG17 & 0.01 \hspace{5mm} (0.02) & 0.33   & 0.50 \hspace{5mm} (0.18)  &  1.90\\ \hline
LOO-RB-RF & 0.01 \hspace{5mm} (0.02) & 0.32 & -0.07 \hspace{5mm} (0.11) & 1.11\\ \hline
Variance LOO-RB-RF & 0.03 \hspace{5mm} (0.00) & 0.07 & -0.21 \hspace{5mm} (0.03) & 0.34 \\ \hline
\end{tabular} \label{tab:other}
\end{table}

When it comes to Total personal expenses, turnover from the tax data is not a good regressor at all, such that the simple GREG1 estimator yields no improvement over the HT estimator. Similarly with the LOO-RB-GREG1 estimator, which performs similarly as the GREG1 estimator, as can be expected. Meanwhile, both the GREG17 and LOO-RB-RF estimators are noticeably better. This suggests that the other regressors can be linearly related to this target variable, and the RF model is flexible enough to automatically capture this linear regression relationship here. The jackknife variance estimators exhibit no bias for the LOO-RB methods in this case.

Turning to the results for Total procurements of goods and services, the LOO-RB-RF estimator is again by far the most efficient of all. The GREG1 and LOO-RB-GREG1 estimators similarly improve on the HT estimator, where turnover from tax data is a reasonable regressor for this variable. The GREG17 estimator is more efficient than the simple GREG1 estimator by a small margin, albeit at the cost of introducing a small bias that is statistically significant. In contrast, the LOO-RB-RF estimator provides a much greater gain of efficiency while remaining design-unbiased. The jackknife variance estimator is essentially unbiased for the LOO-RB-GREG1 estimator, but it has a small negative bias for the LOO-RB-RF estimator.

\subsubsection{Conclusions}

The following conclusions seem warranted based on the simulation results above. 

In situations where simple GREG estimation (with few regressors) have little bias to start with, i.e. when the simple linear regression model is a reasonable statistical model, the proposed unbiased learning approach is unlikely to offer appreciable improvement in practice. More advanced learning techniques cannot be of much help, \emph{without} a supply of additional useful features. Nevertheless, while GREG estimation may suffer from non-negligible bias in a given situation, because the linear model is inappropriate, the unbiased learning approach can avoid the bias automatically. 

More importantly, provided richer auxiliary information, the proposed unbiased SRB learning approach can yield large gains. On the one hand, it allows one to make use of modern ML techniques that can potentially lead to much more flexible and powerful prediction models, without demanding the same kind of effort that is often necessary for building complex parametric models. On the other hand, the theory for design-unbiased statistical learning developed in this paper ensures the resulting ML-assisted estimator is valid for descriptive inference, so that the ML-prediction model can help to generate \emph{valid and efficient} estimation at the aggregated level, wihtout requiring the model to be entirely correct at the individual level, because the prediction errors in the sample are extrapolated to the population of interest based on the known $pq$-sampling design.

\section{Summary remarks} \label{summary}

Amalgamating classic ideas of Statistical Science and Machine Learning, we developed an ML-assisted SRB approach for $pq$-design-unbiased statistical learning in survey sampling. It allows one to generally achieve design-unbiased model-assisted estimation based on probability sampling from the population of interest. The freedom to adopt modern as well as emerging powerful algorithmic ML-prediction models should enable one to make more efficient use of the rich auxiliary information whenever it is available. 

A topic for future research can be noted immediately. As mentioned earlier, it is an open question at this stage how to construct the efficient subsampling scheme $q(s_1 |s)$, including the choice $n_1 = |s_1|$. Moreover, a related issue is the sampling design. In this paper, we have assumed the $pq$-design approach, because it fits naturally with the current practice of survey sampling, where the sampling design $p(s)$ is already implemented and given at the stage of estimation, so that only the subsampling scheme $q(s_1 |s)$ is left to one's own device. However, by construction, the combined randomisation distribution induced by $(p, q)$ is the same as that induced by $(p_1, p_2)$, for any $s_1 \cup s_2 = s$ and $s_1\cap s_2 = \emptyset$. It may be worth investigating whether a direct approach to the design of $(p_1, p_2)$ may offer certain advantages. Finally, it is easily envisaged that more efficient and accurate variance estimation methods will be discovered by future research.


\begin{thebibliography}{99}

\bibitem{blackwell1947} Blackwell, D. (1947). Conditional expectation and unbiased sequential estimation. \textit{Ann. Math. Statist.}, 18: 105-110.

\bibitem{bousquet2002} Bousquet, O. and Elisseeff, A. (2002). Stability and generalization, \textit{J. Mach. Learning Res.}, 2:499-526.

\bibitem{breidt2017} Breidt, F.J. and Opsomer, J.D. (2017). Model-assisted survey estimation with modern prediction techniques. \textit{Statist. Scien.}, 32:190-205.

\bibitem{breiman1996h} Brieman, L. (1996a). Heuristics of instability and stabilization in model selection. \textit{Ann. Statist.}, 24:2350-2383.

\bibitem{breiman1996} Breiman, L. (1996b). Bagging predictors. \textit{Mach. Learn.}, 26:123-140.

\bibitem{cassel1976} Cassel, C. M., S\"{a}rndal, C.-E. and Wretman, J. H. (1976). Some results on generalized difference estimation and generalized regression estimation for finite populations. \textit{Biometrika}, 63:615-620.

\bibitem{deville1992} Deville, J.-C. and S\"{a}rndal, C.-E. (1992). Calibration estimators in survey sampling. \textit{J. Amer. Statist. Assoc.}, 87:376-382.

\bibitem{gordon1978} Gordon, L. and Olshen, R. (1978). Asymptotically Efficient Solutions to the Classification Problem. \textit{Ann.  Statist.}, 6:515-533. 

\bibitem{gordon1980} Gordon, L. and Olshen, R. (1980). Consistent Nonparametric Regression From Recursive Partitioning Schemes. \textit{J. Mult. Ana.}, 10:611-627.

\bibitem{HT1952} Horvitz, D. G. and Thompson, D. J. (1952). A generalization of sampling without replacement from a finite universe. \textit{J. Amer. Statist. Assoc.}, 47:663-685.

\bibitem{mukherjee2006} Mukherjee, S., Niyogi, P., Poggio, T. and Rifkin, R. (2006). Learning theory: stability is sufficient for generalization and necessary and sufficient for consistency of empirical risk minimization. \textit{Adv. Comp. Math.},  25:161-193.

\bibitem{rao1945} Rao, C. R. (1945). Information and accuracy attainable in the estimation of statistical parameters. \textit{Bull. Calcutta Math. Soc.}, 37:81-91.

\bibitem{sarndal2010} S\"{a}rndal, C.-E. (2010). The calibration approach in survey theory and practice. \textit{Surv. Methodol.}, 33:99-119.

\bibitem{sarndal1992} S\"{a}rndal, C.-E., Swensson, B. and Wretman, J. (1992). \textit{Model Assisted Survey Sampling.} New York: Springer-Verlag.

\bibitem{toth2011} Toth, D. and Eltinge, J. L. (2011). Building consistent regression trees from complex sample data. \textit{J. Amer. Statist. Assoc.}, 106:1626-1636. 

\bibitem{tsymbal2004} Tsymbal, A. (2004). The problem of concept drift: definitions and related work. \textit{Comp. Scien.}, 106 (2), 58.

\bibitem{tukey1958} Tukey, J.W. (1958). Bias and confidence in not quite large samples (abstract).  \textit{Ann. Math. Statist.}, 29:614.

\end{thebibliography}
\end{document}